\documentclass[pmlr,twocolumn,10pt]{jmlr}

\usepackage{rotating}
\usepackage{cprotect}
\usepackage{longtable}
\usepackage{booktabs}
\usepackage{siunitx}
\usepackage[switch]{lineno}
\usepackage{longtable}
\usepackage{mathtools}
\usepackage{dsfont}
\usepackage{soul}
\usepackage{algorithm}
\usepackage{pifont}
\usepackage{mdframed}
\usepackage{multirow}
\usepackage{comment}
\usepackage[shortlabels]{enumitem}

\definecolor{mygray}{gray}{0.8}

\newcommand{\cmark}{\ding{51}}
\newcommand{\xmark}{\textcolor{mygray}{\ding{55}}}

\jmlrvolume{}
\jmlrpages{}
\jmlryear{}
\jmlrsubmitted{}
\jmlrpublished{}
\jmlrproceedings{}{}

\title[Representing History for Interpretable Policy Modeling]{How Should We Represent History in Interpretable Models of Clinical Policies?}

\author{%
\Name{Anton Matsson} \Email{antmats@chalmers.se} \\
\addr Chalmers University of Technology and University of Gothenburg
\AND
\Name{Lena Stempfle} \Email{stempfle@chalmers.se} \\
\addr Chalmers University of Technology and University of Gothenburg
\AND
\Name{Yaochen Rao} \Email{yaochenr@student.chalmers.se} \\
\addr Chalmers University of Technology and University of Gothenburg
\AND
\Name{Zachary R. Margolin} \Email{zmargolin@corevitas.com} \\
\addr CorEvitas, LLC
\AND
\Name{Heather J. Litman} \Email{hlitman@corevitas.com} \\
\addr CorEvitas, LLC
\AND
\Name{Fredrik D. Johansson} \Email{fredrik.johansson@chalmers.se} \\
\addr Chalmers University of Technology and University of Gothenburg
}

\begin{document}

\maketitle

\begin{abstract}
Modeling policies for sequential clinical decision-making based on observational data is useful for describing treatment practices, standardizing frequent patterns in treatment, and evaluating alternative policies. For each task, it is essential that the policy model is interpretable. Learning accurate models requires effectively capturing a patient’s state, either through sequence representation learning or carefully crafted summaries of their medical history. While recent work has favored the former, it remains a question as to how histories should best be represented for interpretable policy modeling. Focused on model fit, we systematically compare diverse approaches to summarizing patient history for interpretable modeling of clinical policies across four sequential decision-making tasks. We illustrate differences in the policies learned using various representations by breaking down evaluations by patient subgroups, critical states, and stages of treatment, highlighting challenges specific to common use cases. We find that interpretable sequence models using learned representations perform on par with black-box models across all tasks. Interpretable models using hand-crafted representations perform substantially worse when ignoring history entirely, but are made competitive by incorporating only a few aggregated and recent elements of patient history. The added benefits of using a richer representation are pronounced for subgroups and in specific use cases. This underscores the importance of evaluating policy models in the context of their intended use.
\end{abstract}


\paragraph*{Data and Code Availability.}{We use four medical datasets to model sequential clinical decision-making across various conditions. The Alzheimer's disease dataset comprises 1,605 patients from the Alzheimer’s Disease Neuroimaging Initiative (ADNI) database (\url{https://adni.loni.usc.edu/}). Data on 4,391 patients with rheumatoid arthritis (RA) are sourced from the CorEvitas RA registry~\citep{corevitas}. Sepsis and chronic obstructive pulmonary disease datasets are derived from the MIMIC-III and MIMIC-IV databases~\citep{johnson2016mimic,johnson2023mimic}, containing data on 20,932 and 7,977 patients, respectively. ADNI and MIMIC are publicly available to researchers. RA data are available from CorEvitas, LLC through a commercial subscription agreement and are not publicly available. The code is available at \url{https://github.com/Healthy-AI/inpole}.}

\paragraph*{Institutional Review Board (IRB).}{Research on de-identified data from ADNI and MIMIC is exempt from IRB review under HIPAA. Approval for the investigation of treatment patterns within the CorEvitas RA registry was granted by the Swedish Ethical Review Authority (application no. 2021-06144-01).}

\section{Introduction}
\label{sec:introduction}

Sequential decision-making is central to the treatment of many medical conditions, both acute and chronic~\citep{chakraborty2013statistical,gottesman2019guidelines}. The patterns in how treatment decisions depend on available information, aggregated over physicians and patients, are commonly referred to as the \emph{behavior policy}. Modeling these patterns has several use cases: \textbf{Explanation:} A behavior policy model can provide insights into current treatment strategies and support the development of new clinical guidelines~\citep{pace2022poetree, huyuk2023explaining, deuschel2024contextualized}; \textbf{Implementation:} Identifying common treatment patterns and standardizing them can reduce practice variation and leverage the collective expertise of many clinicians~\citep{esteva2017dermatologist, hannun2019cardiologist}; \textbf{Evaluation:} Most approaches to off-policy evaluation of a new policy, such as importance weighting~\citep{precup2000eligibility}, rely on accurate probabilistic models of the behavior policy.

All three use cases of behavior policy modeling benefit from an \emph{interpretable} model. For explanation and implementation, interpretability is even crucial for gaining the trust of end users~\citep{stiglic2020interpretability}. In off-policy evaluation, interpretability allows for verifying the model fit, reasoning about omitted confounding variables---variables that affect both treatment decisions and outcomes---and comparing the behavior policy to a target policy representing new clinical guidelines~\citep{matsson2022case}. Failing to account for confounders in off-policy evaluation may result in biased estimates of the value of the target policy~\citep{namkoong2020off}. To accurately model sequential clinical decision-making and mitigate bias in downstream tasks, it is important to account for the patient's medical history, including previous contexts, treatments, and outcomes~\citep{gottesman2019guidelines}.
Despite this, several studies neglect historical context, focusing solely on present observations when formulating clinical policies~\citep{javad2019reinforcement,asoh2013application,utomo2018treatment,lin2018deep,weng2017representation}. This raises the question: How should we represent history in interpretable models of clinical policies?

The literature suggests two primary approaches to summarizing patient history: learned sequence representations and hand-crafted summary features. Recent work in interpretable policy modeling favors the former, employing techniques such as recurrent decision trees~\citep{pace2022poetree} or recurrent neural networks~\citep{deuschel2024contextualized} to create abstract history representations. For individual patients, these models provide policy descriptions based on the most recent patient information, individualized through the encoded history. Other representation learning methods identify prototypes~\citep{li2018deep,ming2019interpretable}---patients that represent larger groups of individuals---providing a compact description of the overall policy~\citep{matsson2022case}.

On the other hand, summarizing history using hand-crafted features is useful for fitting simple, interpretable models such as linear or rule-based classifiers. These are arguably more interpretable---at least more transparent~\citep{lipton2018mythos}---than representation learning methods, as they explicitly show how historical information influences the current decision. Naturally, the best summary is specific to the problem at hand~\citep{gottesman2019guidelines}, but several strategies frequently appear in the literature, such as aggregating patient information across time~\citep{raghu2017continuous,komorowski2018artificial,guez2008adaptive}, using a fixed-sized window of the most recent observations~\citep{bertsimas2022data,escandell2014optimization}, or incorporating indicators for past decisions~\citep{bertsimas2022data}. 

Despite the popularity of these approaches to summarizing history, they have not been compared systematically in the context of interpretable policy learning. Notably, prior work using representation learning~\citep{pace2022poetree,deuschel2024contextualized} has not explored the impact of different history representations during evaluation. Moreover, their methods are primarily evaluated on two small datasets, focusing solely on binary decisions.

\paragraph{Contributions.}
We compare diverse methods for representing history in interpretable modeling of clinical policies, asking: How does the quality of the model fit depend on the representation method and the level of detail in history summaries? What factors explain variations in performance across different representation methods? And how does the choice of representation affect common use cases such as explanation, implementation and evaluation? We collect evidence to address these questions by comparing eight history representations across distinct decision-making tasks, featuring both binary and multi-class action spaces: 1) selecting therapy for patients with rheumatoid arthritis, 2) conducting an MRI scan for patients with suspected Alzheimer's disease, and managing 3) sepsis and 4) acute exacerbations of chronic obstructive pulmonary disease in the ICU. In all tasks, interpretable models, whether using representation learning or hand-crafted representations, perform similarly to black-box models in aggregate, suggesting that interpretable policy modeling is viable.
By analyzing patient subgroups, temporal patterns, and critical states, we find that methods with similar average performance differ in the kinds of errors they make and the stages at which these errors occur, which can potentially have major consequences for specific use cases such as off-policy evaluation.

\section{Interpretable Policy Modeling}
\label{sec:problem}

\begin{figure*}[t]
\centering 
\includegraphics[width=0.9\linewidth]{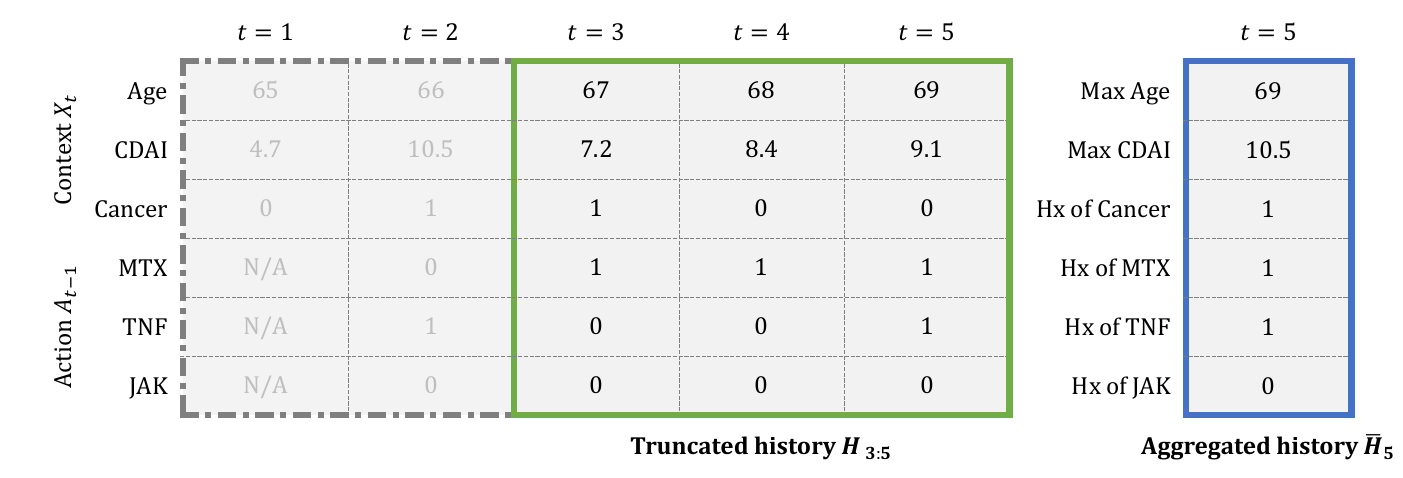}
\cprotect\caption{History truncation and history aggregation using the \verb|max| operator applied to the history of a patient with rheumatoid arthritis. A rolling window of size three is used for the history truncation. The context $X_t$ is a vector with three components, $X_t^1$, $X_t^2$, and $X_t^3$, representing the patient's age, clinical disease activity index (CDAI), and co-existence of cancer. The simplified action space consists of three therapies and their combinations: methotrexate (MTX), tumor necrosis factor (TNF) inhibitor, and Janus kinase (JAK) inhibitor.}
\label{fig:hxrep} 
\end{figure*}

In sequential clinical decision-making, a behavior policy $\mu$ represents the treatment patterns physicians generate when making decisions for patients. Our task is to estimate an unknown behavior policy $\mu$ from observational data consisting of $n$ patient sequences of observations (contexts) $X_t\in\mathcal{X}$ and medical decisions (actions) $A_t\in\mathcal{A}=\{1, \ldots, K\}$, recorded at each stage $t=1, \ldots, T$ of treatment.\footnote{The total number of stages, $T$, is a finite random variable that indicates the end of the course of medication.} We let $H_t \coloneqq (X_1, A_1, \ldots, X_{t-1}, A_{t-1}, X_t)$ represent the history of contexts and actions up until the current stage $t$.\footnote{We assume that the observed outcome of a treatment choice is part of the next set of patient observations.} We refer to the basis for a physician's choice of treatment as the state $S_t\in\mathcal{S}$ of a patient, assumed to be an unknown function of the history $H_t$~\citep{sutton2018rl}. In other words, all direct causes of $A_t$ are assumed to be contained in $S_t$ and in $H_t$. 

We quantify treatment patterns by estimating $p_\mu(A_t \mid S_t)$, the probability of choosing an action $A_t$ in a state $S_t$ under the behavior policy $\mu$. In other contexts, this is called \emph{propensity estimation}~\citep{abadie2016matching}, \emph{policy recovery}~\citep{deuschel2024contextualized} or \emph{behavior cloning}~\citep{torabi2018behavioral}. Particular use cases of estimates $\hat{p}_\mu(A_t \mid S_t)$ may impose additional constraints on the model. To explain sequential clinical decision-making, the model should ideally be fully interpretable, allowing humans to understand its calculations in their entirety. When implementing the policy in clinical practice, it may be sufficient to understand the model's predictions for individual patients to detect errors or unexpected behavior. For off-policy evaluation, the objective is to estimate the importance ratio $\rho$ of a target policy $p_\pi$ and the behavior policy model $\hat{p}_\mu$, $\rho = \frac{p_\pi(A_t \mid S_t)}{\hat{p}_\mu(A_t \mid S_t)}$~\citep{precup2000eligibility}. In this case, an interpretable model of the behavior policy allows for, e.g., understanding differences between the policies and detecting violations of policy overlap~\citep{matsson2022case}.

Following common practice, we construct the state $S_t$ either as a hand-crafted summary or a learned representation of the history, $S_t = f(H_t)$. Regardless of method and use case, a key challenge in policy estimation is to ensure that $S_t$ retains sufficient information to predict the action $A_t$~\citep{gottesman2019guidelines}. This is especially true for interpretable models which aim to have small, transparent policy descriptions. 
In particular, the state must account for confounding variables that have a causal effect on both the treatment decision and its outcome. What constitutes a sufficient state depends on the problem. For example, in the treatment of patients with rheumatoid arthritis, is the choice of treatment $A_t$ based solely on the current context $X_t$, including patient demographics, disease activity measures, and presence of comorbidities? Do previous treatments or their outcomes matter? What about their mutual order? Next, we discuss two common approaches to summarizing a patient's medical history: learned sequence representations and hand-crafted features formed by history truncation and history aggregation. In our experiments, we use hand-crafted features as building blocks to explore different state constructions.

\subsection{Sequence Representation Learning}

Since the space of possible histories $H_t$ grows exponentially with time, the history quickly becomes unwieldy. Sequence models such as recurrent neural networks (RNNs) can be used to learn compact summaries of patient histories to use as the state $S_t$. For example, \citet{wang2018supervised} used a long short-term memory RNN to summarize the history for dynamic treatment recommendation in intensive care. An RNN is known to be opaque but it is possible to open the black-box by introducing a prototype layer into the architecture~\citep{li2018deep,ming2019interpretable} or using it to parameterize an interpretable model~\citep{deuschel2024contextualized}. Another approach is to represent the history using recurrent decision trees~\cite{pace2022poetree}. However, such models require post-processing to enable human interpretation.

\subsection{History Truncation}

History truncation involves selecting a fixed-sized portion of the most recent history, or parts of it, assuming that distant historical events have limited impact on the current decision. Formally, let $H_{(t-k):t} \coloneqq (X_{t-k}, A_{t-k}, \ldots, X_{t-1}, A_{t-1}, X_t)$, where $k \geq 0$, be the truncated history until stage $t$. For $k=2$, as illustrated in Figure~\ref{fig:hxrep}, $H_{(t-2):t}$ includes the current context $X_t$ and contexts from the two preceding stages, along with the actions taken at stage $t-1$, $t-2$ and $t-3$. In our experiments, we apply the history window only to variables for which previous observations are assumed to potentially influence the current decision. For example, in Figure \ref{fig:hxrep}, assuming regular follow-up visits, the patient's age at stage $t-2$ and $t-3$ is redundant given the current age.

Truncating sequence data is a common preprocessing step in natural language processing and bioinformatics. In medical applications, \citet{bertsimas2022data} defined the state based on the most recent heart rate observations to learn a message delivery policy for mobile health. \cite{escandell2014optimization} optimized anemia treatment by formulating a state based on the treatment dose at stage $t-1$, $t-2$ and $t-3$. While truncating history allows for constructing a compact state that captures recent historical events, this representation has two obvious disadvantages. First, truncating the history at a specific time step may exclude critical past information. Second, when $t \leq k$, the absence of earlier history requires some form of imputation, especially if the behavior policy model, $\hat{p}_\mu$, expects an input of fixed size, which is the case for several models in this work.

\subsection{History Aggregation}

History aggregation is applied under the assumption that the temporal order of historical events holds little significance. This method aggregates historical information, such as previous treatment assignments, across time, creating a rough summary of the history. Let $X_t^i$ be a component of the context vector $X_t$. The observations $X_1^i, \ldots, X_t^i$ are combined into a single variable $\bar{X}_t^i$ according to $\bar{X}_t^i = \operatorname*{agg}_t X_t^i$, where the aggregation operator can be, e.g., \verb|sum|, \verb|max| or \verb|mean|. Aggregations of binarized actions are defined analogously, $\bar{A}_t^i = \operatorname*{agg}_t A_t^i$, and the aggregated history is the set of aggregated observations and actions, $\bar{H}_{t}=\{\bar{X}_t, \bar{A}_{t-1}\}$. We apply this operation to variables for which the aggregate is assumed to provide different information than the current observation alone. Again, the age of a patient is an example of a variable for which aggregation provides no extra information. In such cases, we set $\bar{X}_t^i = X_t^i$.

The meaning of history aggregation depends on the aggregation operator and variable type, as illustrated in Figure~\ref{fig:hxrep} using the \verb|max| operator. For numerical variables like CDAI, the aggregate corresponds to the highest observed value. For categorical variables, such as the presence of cancer or previously administered therapies, the aggregate indicates whether the patient has ever experienced the event. Other examples are found in the literature. For example, \citet{komorowski2018artificial} and \citet{raghu2017continuous} accumulated fluids outputs ($\operatorname*{sum}_t X_t^i$)  for learning optimal policies for the management of sepsis. \citet{bertsimas2022data} counted the number of messages previously sent ($\operatorname*{sum}_t A_t^i$) while developing their message delivery policy. \cite{guez2008adaptive} applied \verb|mean| and \verb|max| transformations to EEG signals to optimize stimulation policies for the treatment of epilepsy. 

\begin{table*}[t!]
\small
\centering
\caption{Characteristics of the ADNI, RA, Sepsis, and COPD datasets.}
\label{tab:datasets}
\begin{tabular}{lllll}
\toprule
                                & ADNI              & RA                & Sepsis            & COPD               \\
\midrule
Patients, n                     & 1,605             & 4,391             & 20,932            & 7,977              \\
Age in years, median (IQR)      & 73.9 (69.3, 78.8) & 58.0 (49.0, 66.0) & 66.1 (53.7, 77.9) & 67.0 (56.0, 77.0)  \\
Female, n (\%)                  & 715 (44.5)        & 3355 (76.5)       & 9,250 (44.2)      & 3,472 (43.5)       \\
\midrule
Patient observations $X_t$, n   & 6                 & 33                & 18                & 37                 \\
Actions $A_t$, n                & 2                 & 8                 & 25                & 25                 \\
Stages $T$, median (IQR)        & 3.0 (3.0, 3.0)    & 5.0 (3.0, 8.0)    & 13.0 (10.0, 17.0) & 18.0 (18.0, 18.0)  \\
\bottomrule
\end{tabular}
\end{table*}

\section{Experiments}

We study interpretable models of clinical policies in a series of experiments, aiming to answer the questions raised in Section \ref{sec:introduction}:
How does the quality of the model fit depend on the representation method and the level of detail in the state $S_t$? What factors explain variations in performance across different representation methods? And how does the choice of representation affect common use cases? Recognizing that interpretation is strongly tied to domain knowledge, we do not aim to evaluate the degree of interpretability of the different models. Instead, we seek to understand how the model classes differ in their fit. We compare learning using diverse states based on sequence representation learning and hand-crafted features within decision processes related to four medical conditions: Alzheimer’s disease, rheumatoid arthritis, sepsis, and chronic obstructive pulmonary disease.

\subsection{Datasets}
\label{sec:datasets}

Our datasets, as detailed in the ``Data and Code Availability'' statement, illustrate the diversity of sequential clinical decision-making tasks. For instance, the treatment of Alzheimer’s disease and rheumatoid arthritis (RA) spans several years, with regularly scheduled follow-up visits to slow disease progression. In contrast, managing sepsis and chronic obstructive pulmonary disease (COPD) in the ICU requires continuous administration of treatments like intravenous fluids, vasopressors, and sedative drugs to preserve the patient's life. In all cases except for ADNI, where decisions are binary, clinicians are faced with multiple treatment options at each stage of care. See Table \ref{tab:datasets} for brief characteristics of the datasets. Details are included in Appendix \ref{app:cohorts}.

\subsection{Models}
\label{sec:models}

We include three types of interpretable models based on sequence representation learning: prototype-based models designed for sequential data (PSN)~\citep{ming2019interpretable}, recurrent decision trees (RDT)~\citep{pace2022poetree}, and models leveraging the recent contextualized policy recovery framework (CPR)~\citep{deuschel2024contextualized}. The latter is developed for binary actions and thus only used for ADNI. For comparison, we learn generalized linear models and rule-based models using hand-crafted history representations. Generalized linear models, particularly logistic regression (LR), are widely used as propensity score models for estimating treatment effects in observational studies~\citep{feng2012generalized,spreeuwenberg2010multiple}. Rule-based models, such as decision trees (DT), are commonly employed in clinical decision support systems~\citep{banerjee2019tree,chrimes2023using}. For ADNI, we also include risk scores (RS)~\citep{riskscores}, i.e., scoring systems enabling probabilistic predictions. In addition, we include a multilayer perceptron (MLP) and a recurrent neural network (RNN) in the form of a long short-term memory. These models serve as benchmarks to demonstrate the potential accuracy of policy modeling based on the available data. Table~\ref{tab:models} provides an overview of the included models.

\begin{table*}[t]
\small
\centering
\caption{An overview of the models used in our experiments.}
\label{tab:models}
\begin{tabular}{lccc}
\toprule
Model  & Interpretable policy & Accepts $|\mathcal{A}|>2$ &  Accepts $H_t$  \\
\midrule
Risk scores (RS) & \cmark & \xmark & \xmark \\
Logistic regression (LR) & \cmark & \cmark & \xmark \\
Decision tree (DT) & \cmark & \cmark & \xmark \\
Multilayer perceptron (MLP) & \xmark & \cmark & \xmark \\
\midrule
Contextualized policy recovery (CPR) & \cmark & \xmark & \cmark \\
Prototypical sequence network (PSN) & \cmark & \cmark & \cmark \\
Recurrent decision tree (RDT) & \cmark & \cmark & \cmark \\
Recurrent neural network (RNN) & \xmark & \cmark & \cmark \\
\bottomrule
\end{tabular}
\end{table*}

\subsection{Experimental Setup}
\label{sec:exp_setup}

We divide each dataset---ADNI, RA, Sepsis, and COPD---into training and testing subsets using an 80/20 split, with \SI{20}{\percent} of the training dataset aside for model validation. For the ADNI, RA and COPD datasets, we apply one-hot encoding to categorical features. Depending on the type of model, we standardize continuous variables or discretize them into five equally-sized partitions. Missing values are primarily imputed on a patient level by propagating the last valid observation, secondarily using mean imputation or frequent category imputation. The Sepsis dataset is preprocessed as described in \cite{komorowski2018artificial}, with normally distributed data standardized and log-normally distributed data log-transformed before standardization.

Models based on sequence representation learning are trained used the full history as input, $S_t=H_t$. For other models, we consider the following state representations: $X_t$, $A_{t-1}$, $\{X_t, A_{t-1}\}$, $\bar{H_t}$, $\{X_t, A_{t-1}, \bar{H_t}\}$, $\{H_{(t-1):t}, \bar{H_t}\}$, and $\{H_{(t-2):t}, \bar{H_t}\}$. Note that $\{X_t, A_{t-1}\}=H_{(t-0):t}$. To simplify notation, we let $H_{(k)} \coloneqq H_{(t-k):t} $. For truncation, we compare three operators: \verb|sum|, \verb|max|, and \verb|mean|.

For each dataset, state representation and relevant model type, we train five candidate models using randomly sampled hyperparameters. Final models are selected based on the highest area under the receiver operating characteristic curve (AUROC) score achieved on the validation set. These models are then evaluated on the held-out test set with respect to AUROC and calibration error. The entire process is repeated for five different splits of the dataset, and we report \SI{95}{\percent} confidence intervals based on the bootstrap distribution of the respective metric. Further details on the experimental setup, including hyperparameter selection, can be found in Appendix~\ref{app:exp_details}.

\subsection{General and Stratified Performance}
\label{sec:results}

\begin{table*}[t]
\small
\centering
\cprotect\caption{Average test AUROC, expressed as a percentage, in all four tasks: ADNI, RA, Sepsis, and COPD. The upper section contains models with different hand-crafted states; the lower section contains representation learning methods. MLP and RNN are included as benchmarks. History aggregation $\bar{H}$ is performed using the \verb|sum| operator. Confidence intervals are included in Table \ref{tab:models_full} in Appendix~\ref{app:complexity}.}
\label{tab:results}
\begin{tabular}{llllllllllllllllllllllllll}
\toprule
{} & \multicolumn{4}{c}{\textbf{ADNI}} & \multicolumn{3}{c}{\textbf{RA}} & \multicolumn{3}{c}{\textbf{Sepsis}} & \multicolumn{3}{c}{\textbf{COPD}} \\
\cmidrule(lr){2-5} \cmidrule(lr){6-8} \cmidrule(lr){9-11} \cmidrule(lr){12-14}  
{State} & {RS} & {LR} & {DT} & {MLP} & {LR} & {DT} & {MLP} & {LR} & {DT} & {MLP} & {LR} & {DT} & {MLP} \\
\midrule
$X_t$ & 54.2 & 56.2 & 53.9 & 55.6 & 61.7 & 58.8 & 61.1 & 82.1 & 78.2 & 84.1 & 77.9 & 74.7 & 78.8 \\
$A_{t-1}$ & 52.0 & 53.9 & 53.8 & 53.7 & 94.7 & 94.7 & 94.7 & 88.0 & 90.6 & 91.1 & 92.9 & 95.0 & 95.0 \\
$H_{(0)}$ & 53.4 & 56.8 & 54.3 & 56.8 & 95.6 & 95.7 & 96.1 & 91.3 & 92.1 & 94.7 & 94.0 & 96.0 & 95.4 \\
$\bar{H}_t$ & 63.0 & 64.4 & 64.9 & 64.1 & 90.5 & 92.0 & 94.0 & 84.6 & 85.2 & 89.1 & 91.1 & 89.3 & 93.5 \\
$H_{(0)}, \bar{H}_t$ & 63.7 & 65.3 & 65.0 & 65.8 & 96.1 & 96.5 & 96.9 & 91.9 & 92.3 & 95.3 & 94.7 & 96.7 & 96.3 \\
$H_{(1)}, \bar{H}_t$ & 63.4 & 65.6 & 65.4 & 66.0 & 96.0 & 96.4 & 96.9 & 92.2 & 92.5 & 95.5 & 94.7 & 96.8 & 96.4 \\
$H_{(2)}, \bar{H}_t$ & 62.9 & 65.4 & 65.3 & 66.8 & 96.0 & 96.4 & 96.7 & 92.3 & 92.6 & 95.5 & 94.7 & 96.8 & 96.3 \\
\addlinespace[1.0ex]
\midrule
\addlinespace[1.0ex]
{} & \multicolumn{4}{c}{\textbf{ADNI}} & \multicolumn{3}{c}{\textbf{RA}} & \multicolumn{3}{c}{\textbf{Sepsis}} & \multicolumn{3}{c}{\textbf{COPD}} \\
\cmidrule(lr){2-5} \cmidrule(lr){6-8} \cmidrule(lr){9-11} \cmidrule(lr){12-14}  
{State} & {CPR} & PSN & {RDT} & {RNN} & PSN & {RDT}& {RNN} & PSN & {RDT} & {RNN} & PSN & {RDT} & {RNN} \\
\midrule
$H_t$ & 68.7 & 66.7 & 62.8 & 68.0 & 96.2 & 90.0 & 96.8 & 94.9 & 77.0 & 95.7 & 96.2 & 81.9 & 96.5 \\
\bottomrule
\end{tabular}
\end{table*}

In Table~\ref{tab:results}, we report average test AUROC for each model and dataset, using the different state representations described above, with the \verb|sum| operator used for history aggregation.\footnote{Confidence intervals and calibration errors are included in Table \ref{tab:models_full} and \ref{tab:cal_results}, in Appendix~\ref{app:complexity}. Results for other aggregation operators are shown in Table~\ref{tab:results_mean} and \ref{tab:results_max}.}  Across tasks, we find that the best-performing interpretable model performs on par with RNN, suggesting that interpretable policy modeling is feasible. Accounting for historical information, not just the current context $X_t$, is critical to achieving a good model fit. The best-performing sequence model, which takes the entire history as input, generally outperforms non-sequential models that rely on hand-crafted states. However, the difference in average AUROC is small when providing non-sequential models with a state that captures more than the current context $X_t$ and the previous action $A_{t-1}$. In general, aggregating history through summarization yields the best results. This is particularly true for ADNI, where the difference between aggregation operators is significant. With a well-constructed state, LR can perform significantly better than what is suggested in prior work by, e.g.,~\cite{pace2022poetree}.

In all tasks, models using the state $S_t=A_{t-1}$ perform surprisingly well in terms of average test AUROC. It is reasonable to ask: Is this metric alone a reliable measure of performance? Surely, actions depend on more than the previous choice? To understand this better, we stratify the Sepsis results over treatment stage and by patient groups, defined by the rate of change of the NEWS2 score as in~\citet{luoposition}. We identify six subgroups, enumerated 1--6, where subgroups 1 and 6 correspond to patients that have a large negative and a large positive rate of change of the NEWS2 score, respectively. 

\begin{figure}[t]
\centering 
\includegraphics[width=\linewidth]{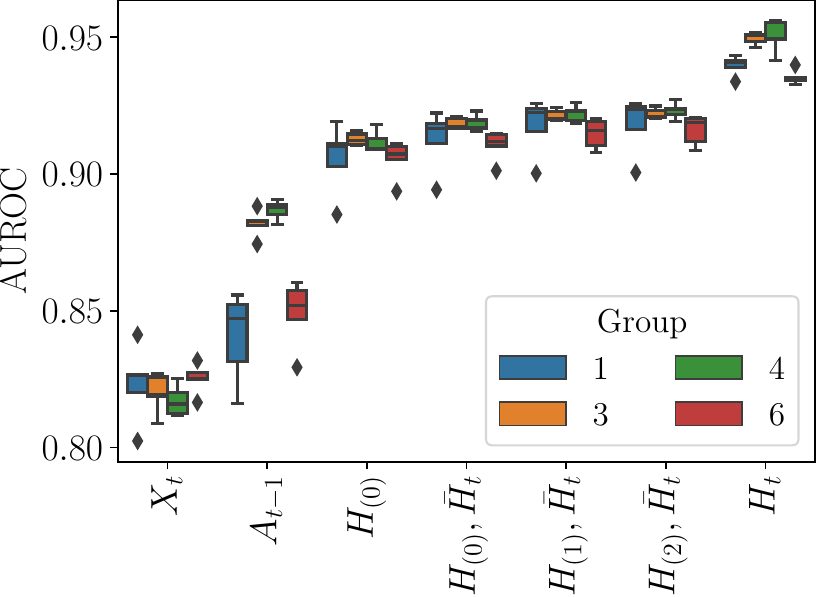}
\cprotect\caption{AUROC across states and patient groups, identified based on the rate of change of the NEWS2 score, in Sepsis. PSN is used with $S_t=H_t$, LR with the others.}
\label{fig:sepsis_groups} 
\end{figure}

\begin{table}[t]%
\centering%
\caption{Percentage difference in AUROC between PSN and LR, fit to the Sepsis data using each state representation in the column ``State''. G1, G3, G4, and G6 denote different patient groups based on the rate of change of the NEWS2 score; G1 and G6 have higher variability than G3 and G4.\label{tab:rel_diffs}}%
\small
\begin{tabular}{l c c c c}
\toprule
State & G1 & G3 & G4 & G6 \\
\midrule
$X_t$ & $-12.5$ & $-13.7$ & $-14.0$ & $-11.8$ \\
$A_{t-1}$ & $-10.5$ & $-7.1$ & $-6.7$ & $-9.2$ \\
$H_{(0)}$ & $-3.6$ & $-3.9$ & $-4.0$ & $-3.2$ \\
$H_{(0)}, \bar{H}_t$ & $-2.9$ & $-3.3$ & $-3.3$ & $-2.6$ \\
$H_{(1)}, \bar{H}_t$ & $-2.4$ & $-3.0$ & $-3.0$ & $-2.2$ \\
$H_{(2)}, \bar{H}_t$ & $-2.3$ & $-2.9$ & $-2.9$ & $-2.1$ \\
\bottomrule
\end{tabular}
\end{table}

In Figure~\ref{fig:sepsis_groups}, we show the distribution of AUROC for subgroups 1, 3, 4, and 6 using all states except $\bar{H}_t$. LR is fit to hand-crafted states, whereas PSN is used with $S_t=H_t$. We clearly see that the previous action is an insufficient state for subgroups 1 and 6, i.e., patient groups that are likely to have higher variation in their treatment compared to other groups. Table~\ref{tab:rel_diffs} clarifies the relative difference in AUROC, expressed as a percentage, between the models. State representations that take historical events into account through truncation and aggregation enable LR to approach the performance of PSN.

\begin{figure}[t]
\centering 
\includegraphics[width=0.9\linewidth]{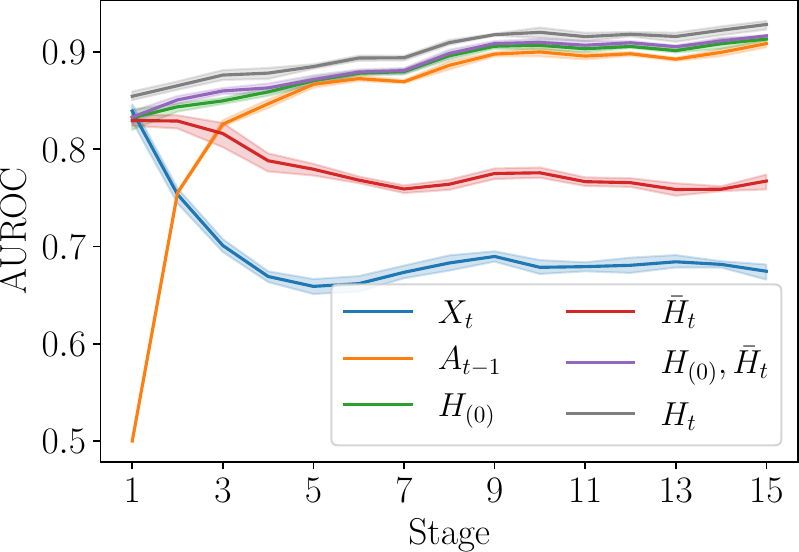}
\cprotect\caption{AUROC for different states across stages of treatment in Sepsis. PSN is used with $S_t=H_t$, DT with the others.}
\label{fig:sepsis_time} 
\end{figure}

Figure~\ref{fig:sepsis_time} shows AUROC with respect to the stage of treatment for DT, fit to the Sepsis data using varying state representations, and PSN with $S_t=H_t$. In early stages, we clearly see the shortcomings of the state based on the previous action only. However, in later stages, $A_{t-1}$ is a good predictor for the doctor's decision. It may be the case that patients' conditions stabilize and that clinicians reuse the same treatment in subsequent stages. The significance of history is illustrated by the difference between $X_t$ and $\bar{H}_t$, with the latter consistently providing a better model fit.

\subsection{Modeling Policies for Explanation, Implementation and Evaluation}

\begin{figure}[t]
\centering 
\includegraphics[width=0.9\linewidth]{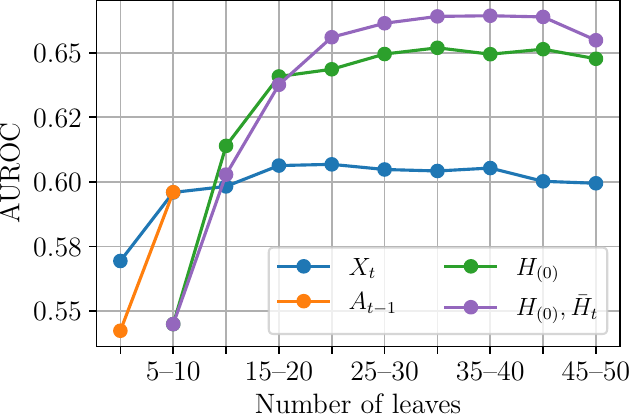}
\cprotect\caption{Therapy switch prediction for RA. AUROC against the number of leaves for DT fit using different states. With $S_t=A_{t-1}$, the tree can have at most 5--10 leaves.}
\label{fig:ra_complexity} 
\end{figure}

Learning an interpretable model of the behavior policy is required to explain decision-making and can help verify assumptions in off-policy evaluation. But is it generally possible to learn an interpretable model that performs well? And is there a cost associated with it? In Figure~\ref{fig:ra_complexity}, we plot AUROC against the number of leaves in decision trees fit to the RA data using four different state representations. We measure AUROC in critical states where a switch of treatment was made. A near-optimal model, obtained with the state $\{H_{(0)}, \bar{H}_t\}$, requires around 30 leaves and may be difficult for humans to comprehend in its entirety, making it less suitable for explanation. However, it could still provide insights into specific decisions, if implemented in practice, as its decision paths (depth) are fairly short. 

\begin{figure}[t]
\centering 
\includegraphics[width=0.9\linewidth]{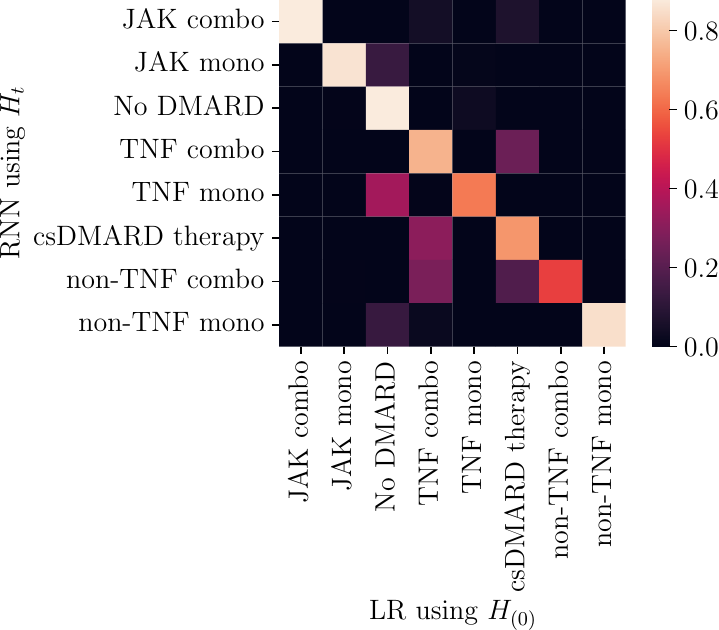}
\cprotect\caption{Errors in therapy switch prediction for RA. Confusion matrix between RNN using $S_t=H_t$ and LR using $S_t=H_{(0)}$. See Appendix~\ref{app:cohorts} for therapy definitions.}
\label{fig:ra_confusion} 
\end{figure}

For implementation, it is logical to ask: How would the actions suggested by a simple model such as LR differ from those suggested by the best-in-class model? In Figure~\ref{fig:ra_confusion}, we investigate this by comparing LR, fit to the RA data using $S_t=H_{(0)}$, to RNN, focusing on the states where a change of treatment was made. In most cases, the predicted actions align, but LR confuses, for example, the less frequent non-TNF combination therapy with the more frequent TNF combination and csDMARD therapies. When the state representation and/or model is overly simplified, we risk losing precision in rare cases.

\begin{figure}[t]
\centering 
\includegraphics[width=0.9\linewidth]{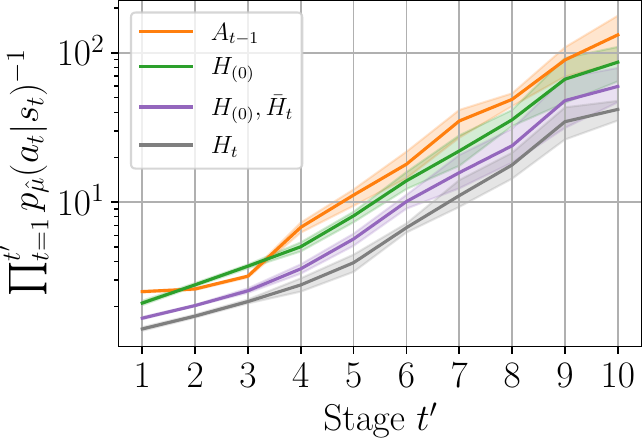}
\cprotect\caption{Off-policy evaluation for RA. Median of inverse probability products at stage $t^{\prime}=1, 2, ..., 10$ of treatment using LR ($A_{t-1}$, $H_{(0)}$, and $\{H_{(0)}, \bar{H}_t\}$) and RNN ($H_t$).}
\label{fig:ra_ope} 
\end{figure}

Off-policy evaluation of new policies is often performed using importance weighting, see Section~\ref{sec:problem}. A crucial step is to re-weight observed outcomes by the product of inverse probabilities $p_{\hat{\mu}}(a_t \mid s_t)^{-1}$, where $a_t$ is the action taken in state $s_t$ under the behavior policy. In Figure~\ref{fig:ra_ope}, we inspect the median of inverse probability products obtained with LR and RNN when considering different state representations for the former. This product is inversely proportional to the likelihood of observed data; the more individual probabilities deviate from 1, the larger the product. The figure shows that using coarse representations of history, such as $A_{t-1}$ or $H_{(0)}$, leads to inverse probability products that grow much faster (note the logarithmic scale) compared to using a more comprehensive representation of history. 
This accelerated growth negatively impacts the variance in off-policy evaluation and poses one of the greatest challenges to its practical feasibility~\citep{gottesman2018evaluating}.

\section{Discussion}
\label{sec:discussion}

In this work, we compared two common approaches to representing patient history for interpretable modeling of clinical policies: hand-crafted summary features and learned sequence representations. In particular, we studied how the quality of the model fit depends on the representation method and the level of detail in history summaries. Across four decision-making tasks, we found it possible to achieve competitive results using simple, manually crafted representations. Combining current patient observations, the most recent treatment, and historical aggregates of prior observations and treatments explained most of the variance in treatment selection. Notably, incorporating recent treatments was critical to model performance. These findings are consistent with clinical guidelines. For example, current recommendations for the management of rheumatoid arthritis focus on broad indicators such as poor prognostic factors rather than details of the patient's medical history~\citep{eular}. 

We investigated factors that explain variations in performance across different representation methods. For instance, by breaking down the results by patient subgroups and stages of treatment, we were able to identify shortcomings in simplified representations. Additionally, focusing on therapy selection in rheumatoid arthritis, we highlighted challenges associated with common use cases of interpretable policy modeling. For example, using a coarse history may increase variance in off-policy evaluation. In all experiments, interpretable models using learned representations performed comparably to black-box models, suggesting that interpretable policy learning is generally viable.

We assumed that all direct causes of treatment selections were captured within the observed patient histories. In practice, we were limited to the variables that were actually measured, which means there could be unmeasured confounders. In ADNI, we observed an average AUROC of 0.6--0.7, suggesting that variance in MRI scan ordering is not fully explained by the available variables. However, the published diagnostic policy for mild cognitive impairment, which can be modeled with the variables used here, shows no clear evidence of omitted variables~\citep{pace2022poetree}. The variance may instead stem from differences across institutions and practitioners.

Another limitation is that we examined only a limited set of manually created history summaries. For example, it would be possible to combine different aggregation methods or derive additional features from historical data, such as changes in critical observations over time. We leave this extension for future research. The primary goal of this work was to understand the overarching impact of historical information in policy modeling, rather than to fine-tune representations for individual tasks.

An interesting direction for future work is to construct policy models that explicitly depend on the stage of treatment. As shown in Figure 2, current patient observations are important for accurately predicting the initial treatments of sepsis. Later in the process, the treatment is often repeated, suggesting that the patients' conditions stabilize. However, although a simple model may be sufficient to explain overall patterns, it risks introducing severe bias in specific use cases such as policy evaluation.

\acks{
This work was partially supported by the Wallenberg AI, Autonomous Systems and Software Program
(WASP) funded by the Knut and Alice Wallenberg Foundation.

The computations and data handling were enabled by resources provided by the National Academic Infrastructure for Supercomputing in Sweden (NAISS), partially funded by the Swedish Research Council through grant agreement no. 2022-06725.
}

\bibliography{references}

\appendix

\section{Dataset Descriptions}
\label{app:cohorts}

In this section, we describe the datasets used in our experiments (ADNI, RA, Sepsis, and COPD) in more detail. When using a state representation based on history truncation, we replaced missing context observations for $t \leq k$ with the corresponding observation at the first time step. Missing information about the previous action(s) was replaced with ``csDMARD therapy'' (RA), ``no MRI scan'' (ADNI), and 0 (Sepsis and COPD).

\subsection{Alzheimer’s Disease}

Following \cite{huyuk2023explaining} and \cite{pace2022poetree}, we compiled a dataset of 1,605 patients from the Alzheimer’s Disease Neuroimaging Initiative (ADNI) database (\url{https://adni.loni.usc.edu/}). The raw data was loaded using the function \verb|adnimerge| provided in the R package ``ADNIMERGE'', see \url{https://adni.bitbucket.io}. Specifically, we filtered out visits lacking a CDR-SB measurement, visits with a separation of more than six months, and patients with fewer than two visits.\footnote{The sum of boxes of the clinical dementia rating (CDR-SB) is a measure of dementia.} We estimated the behavior policy for determining whether patients with suspected cognitive impairment should undergo a magnetic resonance imaging (MRI) scan. Patient observations included the CDR-SB score as well as factors such as age, gender, marital status, education level, and possession of the APOE4 allele. Following \cite{huyuk2023explaining} and \cite{pace2022poetree}, we categorized the CDR-SB score as normal (0--0.5), questionable (0.5--4.5) or severe (4.5--18.0). The previous action was encoded in the outcome of any MRI scan ordered at the previous visit (no MRI scan ordered; hippocampal volume below average, average, or above average). Hippocampus volumes that deviated by  $\pm0.5$ standard deviations of the average hippocampus volume were classified as below and above average, respectively. We describe the context variables in Table \ref{tab:adni_statics}.

\begin{table*}[t!]
\small
\centering
\caption{A summary of the variables included in the context vector $X_t$ in the ADNI experiments and their baseline statistics. We applied history aggregation and history truncation only to the variables in the lower section of the table. For each variable, N represents the number of patients with non-missing baseline information.}
\label{tab:adni_statics}
\begin{tabular}{lll}
\toprule
Variable & N & Statistics \\
\midrule
Age in years, median (IQR) & 1605 & 73.9 (69.3, 78.8) \\
Education level, median (IQR) & 1605 & 16 (14, 18) \\
Female, n (\%) & 1605 & 715 (44.5) \\
Marital status, n (\%) & 1605 & \\
\qquad Married && 1217 (75.8) \\
\qquad Divorced && 134 (8.3) \\
\qquad Widowed && 191 (11.9) \\
\qquad Never married &&  56	(3.5) \\
\qquad Unknown && 7 (0.4) \\
APOE4 alleles, n (\%) & 1605 & \\
\qquad 0 && 840 (52.3) \\
\qquad 1 && 600 (37.4) \\
\qquad 2 && 165	(10.3) \\
\midrule
CDR-SB, n (\%) & 1605 & \\
\qquad Normal && 453 (28.2) \\
\qquad Questionable && 985	(61.4) \\
\qquad Severe && 167 (10.4) \\
\bottomrule
\end{tabular}
\end{table*}

\subsection{Rheumatoid Arthritis}
\label{subsec:ra_details}

We used data from the CorEvitas RA registry \citep{corevitas}, an ongoing longitudinal clinical registry in the US, to model the behavior policy for choosing disease-modifying antirheumatic drug (DMARD) therapy for patients with RA. The standard procedure in treating these patients involves initiating a conventional synthetic DMARD (csDMARD) and incorporating a biologic or targeted synthetic DMARD (b/tsDMARD) if the initial therapy fails, see for example \cite{eular}. Biologic DMARDs are commonly divided into Tumor necrosis factor (TNF) biologics and non-TNF biologics. Janus kinase inhibitors (JAKi) are the most frequently used tsDMARDs.

The raw dataset contained data from 42,068 patients enrolled in the registry between January 2012 and December 2021. We removed 1,323 patients for whom information on therapy changes were missing or apparently incorrect. Since patients may have joined the registry at different times in their disease course, we focused on subsequences starting with the initiation of the first b/tsDMARD. We excluded patients with a history of b/tsDMARD treatment at registry enrollment and patients who did not start any b/tsDMARD within the registry. Additionally, we excluded all visits, as well as any subsequent visits, where multiple b/tsDMARDs were prescribed, as this is not clinically recommended. This left us with 4,391 patients in the final cohort. 

There were no constraints placed on the follow-up visits in terms of regularity, although the registry protocol suggests visits every six months in line with clinical practice. Each patient was monitored up until their final documented visit or the data cut date of December 31, 2021, whichever came earlier. Consequently, the total number of registry visits and the length of follow-up differed among patients. We included 33 variables, see Table \ref{tab:ra_statics1} and \ref{tab:ra_statics2}, in the context vector $X_t$.

Following previous work by \cite{matsson2023patterns}, we studied changes between classes of DMARDs rather than changes between individual DMARDs. The following classes of drugs were studied: csDMARD therapy (``csDMARD''), TNF biologic monotherapy (``TNF (m)''), TNF biologic combination therapy (``TNF (c)''), non-TNF biologic monotherapy (``non-TNF (m)''), non-TNF biologic combination therapy (``non-TNF (c)''), JAKi monotherapy (``JAK (m)''), JAKi combination therapy (``JAK (c)''), and no DMARD therapy (``No DMARD'').

\begin{table*}[t!]
\small
\centering
\caption{A  summary of variables included in the context vector $X_t$ in the RA experiment and their baseline statistics. We applied neither history aggregation nor history truncation to these variables. The other context variables, for which we applied these operations, are listed in Table \ref{tab:ra_statics2}. Baseline refers to the second visit, i.e., the first visit for which we could determine the switch label. For each variable, N represents the number of patients with non-missing baseline information.}
\label{tab:ra_statics1}
\begin{tabular}{lll}
\toprule
Variable & N & Statistics \\
\midrule
Age in years, median (IQR) & 4379 & 59 (50, 67) \\
Calendar year, median (IQR) & 4391 & 2016 (2013, 2019) \\
RA duration in years, median (IQR) & 4338 & 3 (1, 9) \\
Female, n (\%) & 4383 & 3355 (76.5) \\
College completed, n (\%) & 3997 & 1507	(37.7) \\
Work status, n (\%) & 4291 & \\ 
\qquad Full time && 1784 (41.6) \\
\qquad Part time && 359 (8.4) \\
\qquad Work at home && 363 (8.5) \\
\qquad Student && 68 (1.6) \\
\qquad Disabled && 450 (10.5) \\
\qquad Retired && 1267 (29.5) \\
 Private insurance, n (\%) &4391& 3128 (71.2) \\
 Medicare insurance, n (\%) &4391& 1412 (32.2) \\
 Medicaid insurance, n (\%) &4391& 278 (6.3) \\
 No insurance, n (\%) &4391& 95	(2.2) \\
\bottomrule
\end{tabular}
\end{table*}

\begin{table*}[t!]
\small
\centering
\caption{A summary of variables included in the context vector $X_t$ in the RA experiment and their baseline statistics. We applied both history aggregation and history truncation to these variables. The other context variables, for which we did not apply these operations, are listed in Table \ref{tab:ra_statics1}. Baseline refers to the second visit, i.e., the first visit for which we could determine the switch label. For each variable, N represents the number of patients with non-missing baseline information.}
\label{tab:ra_statics2}
\begin{tabular}{lll}
\toprule
Variable & N & Statistics \\
\midrule
BMI, n (\%) & 4304 & \\ 
\qquad Underweight && 49 (1.1) \\
\qquad Healthy && 965 (22.4) \\
\qquad Overweight && 1313 (30.5) \\
\qquad Obesity && 1977 (45.9) \\
Blood pressure, n (\%) & 4321 & \\ 
\qquad Normal && 1073 (24.8) \\
\qquad Elevated && 660 (15.3) \\
\qquad Hypertension stage 1 && 1407 (32.6) \\
\qquad Hypertension stage 2 && 1154 (26.7) \\
\qquad Hypertension stage 3 && 27 (0.6) \\
CDAI, n (\%) & 4308 & \\ 
\qquad Remission && 489 (11.4) \\
\qquad Low && 1025 (23.8) \\
\qquad Moderate && 1405	(32.6) \\
\qquad High && 1389	(32.2) \\
Smoker, n (\%) & 3757 & 644 (17.1) \\
Drinker, n (\%) & 4281 & 1934 (45.2) \\
Currently pregnant, n (\%) & 2982 & 4 (0.1) \\
Pregnant since last visit, n (\%) & 2418 & 13 (0.5) \\
CCP positive, n (\%) &721& 407 (56.4) \\
RF positive, n (\%) &791 & 499 (63.1) \\
PPD positive, n (\%) &780 & 45 (5.8) \\
Erosive disease, n (\%) &3403 & 253	(7.4) \\
Joint space narrowing, n (\%) &985 & 541 (54.9) \\
Joint deformity , n (\%) &965 & 154	(16.0) \\
Comorbidities, n (\%) & & \\
\qquad Severe infections & 4391 & 66 (1.5) \\
\qquad Metabolic diseases & 4390 & 312 (7.1) \\
\qquad Cardiovascular diseases & 4390 & 486	(11.1) \\
\qquad Respiratory diseases & 4353 & 112 (2.6) \\
\qquad Cancer & 4390 & 125 (2.8) \\
\qquad GI and liver diseases &4390 & 60	(1.4) \\
\qquad Musculoskeletal disorders & 4340 & 1296 (29.9) \\
\qquad Other diseases & 4390 & 516 (11.8) \\
\bottomrule
\end{tabular}
\end{table*}

\subsection{Sepsis}

Utilizing data from the MIMIC-III database \citep{johnson2016mimic}, we modeled the behavior policy for administering vasopressors and intravenous (IV) fluids to patients diagnosed with sepsis in the ICU. The data, comprising 20,932 patients, was structured as multivariate time series with a discrete time step of four hours using code provided by \cite{komorowski2018artificial}. The patients were followed for up to 72 hours, but some individual sequences were shorter as a result of discharge or death of the patient. The doses of vasopressors and IV fluids were discretized into 5 distinct levels and then combined to form a 25-dimensional action space. We included a subset of the available features, see Table \ref{tab:sepsis_statics} for details.  We represented the previous action in terms of the actual doses of vasopressors and IV fluids in the previous 4-h period.

To highlight differences between models and state representations, we stratified the Sepsis results by patient subgroups according to the average rate of change of the National Early Warning Score 2 (NEWS2) score~\citep{inada2018news}. Following \cite{luoposition}, we used the following intervals for the average rate of change of the NEWS2 score (x): $x<-0.4$; $-0.4\leq x<-0.15$; $-0.15\leq x < 0$; $0 \leq x < 0.15$, $0.15 \leq x < 0.4$, and $x>0.4$. In contrast to \cite{luoposition}, we did not consider the variance in $x$ to further separate the patients within each group.

\begin{table*}[t!]
\small
\centering
\caption{A summary of the variables included in the context vector $X_t$ in the sepsis experiment and their baseline statistics. We applied history aggregation and history truncation only to the variables in the lower section of the table. For each variable, N represents the number of patients with non-missing baseline information.}
\label{tab:sepsis_statics}
\begin{tabular}{lll}
\toprule
Variable & N & Statistics \\
\midrule
Age in years, median (IQR) & 20932 & 66.1 (53.7, 77.9) \\
Female, n (\%) & 20932 & 9250 (44.2) \\
\midrule
Heart rate, median (IQR) & 20932 & 87.2 (75.7, 99.8) \\
SysBP, median (IQR) & 20932 & 118.2 (105.4, 133.8) \\
DiaBP, median (IQR) & 20932 & 56.8 (48.2, 66.0) \\
MeanBP, median (IQR) & 20932 & 77.2 (69.0, 87.2) \\
Shock index, median (IQR) & 20932 & 0.7 (0.6, 0.9) \\
Hemoglobin, median (IQR) & 20932 & 10.5 (9.3, 12.1) \\
Blood urea nitrogen, median (IQR) & 20932 & 22.9 (15.0, 37.3) \\
Creatinine, median (IQR) & 20932 & 1.0 (0.8, 1.6) \\
Total urine output, median (IQR) & 20932 & 0.0 (0.0, 250.0) \\
Base excess, median (IQR) & 20932 & 0.0 ($-2.2$, 3.0) \\
Lactate, median (IQR) & 20932 & 1.7 (1.2, 2.5) \\
pH, median (IQR) & 20932 & 7.4 (7.3, 7.4) \\
HCO3, median (IQR) & 20932 & 24.3 (21.0, 27.4) \\
$\text{PaO}_{\text{2}}/\text{FiO}_{\text{2}}$ ratio, median (IQR) & 20932 & 265.0 (173.3, 417.5) \\
Elixhauser, median (IQR) & 20932 & 4.0 (2.0, 5.0) \\
SOFA, median (IQR) & 20932 & 7.0 (5.0, 9.0) \\
\bottomrule
\end{tabular}
\end{table*}

\subsection{Chronic Obstructive Pulmonary Disease}

To extract the COPD dataset, we executed the \verb|main| script in the MIMIC-IV pipeline provided by \cite{gupta2022extensive}, focusing on ICU patients diagnosed with COPD as the chronic disease of interest, totaling 8,535 patients. Initially, we incorporated data on diagnoses, procedures, medications, outputs, and chart events. Subsequently, we conducted a clinical grouping of diagnoses based on their medical codes to streamline the feature space. In addition, ICD-9 codes were mapped to ICD-10 codes in cases where both coding systems were used. The preprocessing of the COPD dataset further entailed the cleaning of lab and chart events through outlier removal and unit conversion, resulting in a final dataset of 7,977 patients. Outlier removal aimed to eliminate values that exceeded the 0.75 percentile threshold and those that fell below the 0.25 percentile threshold across all values for each time point. 

We formatted the data collected over 72 hours as multidimensional time series with a discrete time step of 4 hours, resulting in 18 observations for each patient. We investigated the behavior policy for managing IV fluids (mainly Dextrose 5\% and NaCl 0.9\%) and sedative drugs, combined into 25 discrete actions. Sedative drugs for induction and maintenance of general anesthesia included propofol and fentanyl concentrate. The previous action was represented through the actual doses of IV fluids and sedatives in the previous 4-h period. Missing values were zero-imputed. In cases where values were recorded at different times, forward and backward imputation were applied to minimize bias. We removed columns that had more than \SI{80}{\percent} missingness. The final cohort comprised demographic data and chart events, representing records on vital signs such as blood pressure, heart rate, respiratory rate, and body temperature. We included 40 variables in the context vector $X_t$, as detailed in Table~\ref{tab:dataset_statics_copd1} and \ref{tab:dataset_statics_copd2}.

\begin{table*}[!t]
\small
\centering
\caption{A summary of the variables included in the context vector $X_t$ in the COPD experiments and their baseline statistics. We applied neither history aggregation nor history truncation to these variables. The other context variables, for which we applied these operations, are listed in Table \ref{tab:dataset_statics_copd2}. For each variable, N represents the number of patients with non-missing baseline information.}
\label{tab:dataset_statics_copd1}
\begin{tabular}{lll}
\toprule
Variable & N & Statistics \\
\midrule
Age in years, median (IQR) & 7977 & 67.0 (56.0, 77.0) \\
Female, n (\%) & 7977 & 3472 (43.52) \\
Ethnicity (self-reported), n (\%) & 7977 & \\
\quad White && 5417 (67.91) \\
\quad Black/African American && 747 (9.36) \\
\quad Hispanic/Latino && 242 (3.03)\\
\quad Asian && 203 (2.54)\\
\quad American Indian/Alaska Native && 16 (0.20)\\
\quad Other && 375 (4.70) \\
\quad Unknown/Unable to obtain && 977 (12.24)\\
Health insurance, n (\%) &7977& \\ 
\quad Medicare && 4066 (50.97)\\
\quad Medicaid && 522 (6.54)\\
\quad Other && 3389 (42.48) \\
\bottomrule
\end{tabular}
\end{table*}

\begin{table*}[!t]
\small
\centering
\caption{A summary of the variables included in the context vector $X_t$ in the COPD experiment and their baseline statistics. We applied both history aggregation and history truncation to these variables. The other context variables, for which we did not apply these operations, are listed in Table \ref{tab:dataset_statics_copd1}. For each variable, N represents the number of patients with non-missing baseline information. We present the original naming of the features from the MIMIC database for reproducibility.}
\label{tab:dataset_statics_copd2}
\begin{tabular}{lll}
\toprule
Variable & N & Statistics \\
\midrule
CHART 220045, median (IQR) & 7958 &  84.5 (72.6, 97.3) \\
CHART 220046, median (IQR) & 7954 &  120.0 (30.0, 130.0)\\
CHART 220047, median (IQR) & 7956 &  50.0 (12.5, 30.0)  \\
CHART 220179, median (IQR) & 7629 &  108.5 (91.3,125.0) \\
CHART 220180, median (IQR) & 7627 &  58.0 (46.0, 68.5) \\
CHART 220181, median (IQR) & 7631 &  71.0 (59.0, 82.0)\\
CHART 220210, median (IQR) & 7957 &  19.3 (16.3, 22.5) \\
CHART 220228, median (IQR) & 7899 &  10.5 (0.0, 10.5)  \\
CHART 220277, median (IQR) & 7959 &  96.8 (94.8, 98.5)\\ 
CHART 220545, median (IQR) & 7899 &  25.7 (0.0, 31.6) \\ 
CHART 220546, median (IQR) & 7949 &  7.5 (0.0, 12.3) \\
CHART 220602, median (IQR) & 7955 &  99.0 (21.0, 106.0)\\
CHART 220615, median (IQR) & 7930 &  0.7 (0.0, 1.2)\\ 
CHART 220621, median (IQR) & 7971 &  101.0 (0.0, 136.0)\\ 
CHART 220635, median (IQR) & 7946 &  1.8 (0.0, 2.1)\\
CHART 220645, median (IQR) & 7964 &  135.0 (31.5, 140.0)\\
CHART 223751, median (IQR) & 7558 &  160.0 (0.0, 140.0)  \\
CHART 223752, median (IQR) & 7552 &  90.0 (0.0, 90.0) \\
CHART 223769, median (IQR) & 7956 &  100.0 (25.0, 100.0) \\
CHART 223770, median (IQR) & 7956 &  90.0 (92.0, 22.5)\\
CHART 224161, median (IQR) & 7957 &  30.0 (7.5, 35.0) \\
CHART 224162, median (IQR) & 7955 &  8.0 (2.0, 8.0) \\ 
CHART 225624, median (IQR) & 7935 &  13.0 (0.0, 26.0) \\
CHART 225625, median (IQR) & 7939 &  7.8 (0.0, 8.5) \\
CHART 225677, median (IQR) & 7941 &  2.6 (0.0, 3.7)\\
CHART 226253, median (IQR) & 7947 &  85.0 (21.3, 86.5) \\
CHART 227073, median (IQR) & 7969 &  11.0 (0.0, 15.0) \\
CHART 227442, median (IQR) & 7971 &  3.7 (0.8, 4.2)\\
CHART 227443, median (IQR) & 7941 &  20.0 (0.0, 24.0) \\
CHART 227457, median (IQR) & 7942 &  110.0 (0.0, 204.0) \\
CHART 227465, median (IQR) & 7573 &  12.0 (0.0, 14.6) \\ 
CHART 227466, median (IQR) & 7563 &  26.3 (0.0, 33.2) \\
CHART 223761, median (IQR) & 7722 &  97.7 (24.6, 101.5)\\
\bottomrule
\end{tabular}
\end{table*}

\section{Experimental Details}
\label{app:exp_details}

The results presented in Table \ref{tab:results} were obtained by, for each dataset, state representation, and model class, averaging the AUROC of the \textit{best performing} model across five different splits of the dataset. Specifically, for each such setting, five candidate models were trained using hyperparameters randomly sampled from predefined distributions, and the best performing model was selected as the one with the highest AUROC (ADNI and RA) or accuracy (Sepsis and COPD) on a held-out validation set comprising \SI{20}{\percent} of the training data. The hyperparameter distributions for each model are shown in Table \ref{tab:general_hyperparameters} and \ref{tab:experiment_dependent_hp}. Models based on the contextualized policy recovery framework are not included in the table as we trained them using the code provided by \cite{deuschel2024contextualized}, using their hyperparameter ranges. We refer to their work for details.

The neural networks, including the prototype-based models and the recurrent decision tree, were implemented using PyTorch~\citep{paszke2019pytorch} in combination with skorch~\citep{skorch} and trained on Nvidia Tesla T4 GPUs. Model parameters were optimized using the cross-entropy loss and the Adam optimizer with default parameters. ReLU and hyperbolic tangent were used as activation function in feedforward and recurrent neural networks, respectively. Early stopping was applied to the training if there was no improvement in performance for 5 (ADNI and RA) or 25 (Sepsis and COPD) consecutive epochs. The logistic regression and the decision tree classifier were implemented using the scikit-learn library~\citep{pedregosa2011scikit}. For the scoring system, we used the implementation provided in \cite{riskscores}. The computational time required to produce the results presented in this paper was approximately 2000 core-hours.

For the prototype-based models, three regularization terms, encouraging diversity, clustering, and evidence, were added to the loss function, see \cite{ming2019interpretable} for details. We set the parameters for clustering and evidence regularization to \SI{0.001} and sampled parameter values for diversity regularization, see Table \ref{tab:general_hyperparameters}. Following \cite{matsson2022case}, each learned prototype $i$ was a subsequence of length $t_{i} \leq T$ of a patient sequence in the training data. Prototype projections, see Equation (6) in \cite{matsson2022case}, were performed every fifth epoch. The recurrent decision tree was implemented as in \cite{pace2022poetree}, using the predictive distribution from the leaf with the greatest path probability. We found that the post-processing steps in \cite{pace2022poetree} drastically reduced the performance of the model; hence, we evaluated the recurrent decision trees without applying any post-processing steps.

Figure~\ref{fig:ra_complexity} shows how the performance of decision trees varies with their complexity, measured by the number of leaves, for different state representations in RA. Since we could not control the number of leaves directly, we trained 500 different models for each state representation, using randomly selected hyperparameters. We then binned the models based on their complexity and selected the best-performing model in each ``complexity bucket'' (e.g., 10--20 leaves) to present in the figure. We only performed this experiment for a single split of the data.

\begin{table*}[!t]
\small
\centering
\caption{Experiment-independent hyperparameters and their respective search space for all models.}
\label{tab:general_hyperparameters}
\begin{tabular}{lll}
\toprule
Model & Hyperparameter & Search space \\
\midrule
\multirow{3}{*}{RS} 
& max coefficient value  & $\{3, 4, 5, 6, 7, 8\}$ \\
& max model size  & $\{3, 4, 5, 6, 7\}$ \\
& positive class weight & $\{1, 2, 3, 4, 5\}$ \\
\midrule
\multirow{3}{*}{LR} 
& penalty & \{L2\} \\
& C & $\{10^{-3}, 10^{-2}, 10^{-1}, 10^{0}, 10^{1}, 10^{2}, 10^{3}\}$ \\
& max iterations & $\{2000\}$ \\
\midrule
\multirow{2}{*}{DT} 
& criterion & \{gini, entropy\} \\
& min samples to split & $\{2, 4, 8, 16, 32, 64, 128\}$ \\
\midrule
\multirow{2}{*}{MLP} 
& hidden dimensions (encoder) & $\{(16,), (32,), (64,), (16, 16), (32, 32), (64, 64)\}$ \\
& output dimensions (encoder) & $\{16, 32, 64\}$ \\
\midrule
\multirow{4}{*}{PSN} 
& prototype threshold $d_{min}$ & $\{1, 2, 3, 4, 5\}$ \\
& diversity regularization $\lambda_d$ & $\{10^{-5}, 10^{-4}, 10^{-3}, 10^{-2}, 10^{-1}, 10^{0}\}$ \\
& output dimensions (encoder) & $\{16, 32, 64\}$ \\
& number of layers (encoder) & $\{1, 2\}$ \\
\midrule
\multirow{7}{*}{RDT} 
& initial depth & $\{1, 2\}$ \\
& splitting penalty $\lambda$ & $\{-3, -2, -1\}$ \\
& max depth & $\{3, 4, 5\}$ \\
& evolution prediction $\delta_1$ & $\{10^{-3}, 10^{-2}, 10^{-1}\}$ \\
& evolution prediction $\delta_2$ & $\{10^{-3},  10^{-2},  10^{-1}\}$ \\
& history dimension & $\{5, 10, 15, 20\}$ \\
\midrule
\multirow{2}{*}{RNN} 
& output dimensions (encoder) & $\{16, 32, 64\}$ \\
& number of layers (encoder) & $\{1, 2\}$ \\
\bottomrule
\end{tabular}
\end{table*}

\begin{table*}[!t]
\small
\centering 
\caption{Experiment-dependent hyperparameters of DT and the neural network-based models, along with their search space on different datasets.}
\label{tab:experiment_dependent_hp}
\begin{tabular}{lllll}
\toprule
Model & Hyperparameter & ADNI & RA & Sepsis/COPD \\
\midrule
\multirow{1}{*}{DT} 
& max depth & $\{3, 5, 7, 9, 11, 13, 15\}$ & $\{2, 3, 4, 5, 6, 7, 8\}$ & $\{3, 5, 7, 9, 11, 13, 15\}$\\
\midrule
\multirow{3}{*}{MLP} 
& learning rate & $\{10^{-3}, 10^{-2}\}$ & $\{10^{-3}, 10^{-2}\}$ & $\{10^{-4}, 10^{-3}, 10^{-2}\}$ \\
& max epochs & \{20\} & \{50\} & \{500\}\\
& batch size & $\{16, 32, 64\}$ & $\{128, 256\}$ & $\{256, 512, 1024\}$\\
\midrule
\multirow{3}{*}{PSN} 
& learning rate & $\{10^{-3}, 10^{-2}\}$ & $\{10^{-3}, 10^{-2}\}$ & $\{10^{-4}, 10^{-3}, 10^{-2}\}$ \\
& max epochs & \{20\} & \{50\} & \{500\}\\
& batch size & $\{16, 32, 64\}$ & $\{32, 64\}$ & $\{16, 32, 64\}$ \\
& number of prototypes & $\{2, 4, 6, 8, 10\}$ & $\{2, 4, 6, 8, 10\}$ & $\{5, 10, 15, 20, 25, 30\}$ \\
\midrule
\multirow{3}{*}{RDT} 
& learning rate & $\{10^{-3}, 10^{-2}\}$ & $\{10^{-3}, 10^{-2}\}$ & $\{10^{-4}, 10^{-3}, 10^{-2}\}$ \\
& max epochs & \{20\} & \{50\} & \{500\}\\
& batch size & $\{16, 32, 64\}$ & $\{32, 64\}$ & $\{16, 32, 64\}$\\
\midrule
\multirow{3}{*}{RNN} 
& learning rate & $\{10^{-3}, 10^{-2}\}$ & $\{10^{-3}, 10^{-2}\}$ & $\{10^{-4}, 10^{-3}, 10^{-2}\}$ \\
& max epochs & \{20\} & \{50\} & \{500\}\\
& batch size & $\{16, 32, 64\}$ & $\{32, 64\}$ & $\{16, 32, 64\}$ \\
\bottomrule
\end{tabular}
\end{table*}

\section{Supplementary Results}
\label{app:complexity}

In Table \ref{tab:cal_results}, we show the average test calibration error for each model and dataset, using the different state representations described in Section \ref{sec:exp_setup}. We consider the expected calibration error (ECE) and the static calibration error (SCE)~\citep{nixon2019calibration}, a multi-class extension of ECE. Specifically, we report ECE for ADNI and SCE for RA, Sepsis, and COPD. In Table \ref{tab:models_full}, we report \SI{95}{\percent} confidence intervals, based on the bootstrap distribution of the average AUROC, for the results in Table \ref{tab:results}. Table~\ref{tab:results_mean} and Table~\ref{tab:results_max} show the average test AUROC with history aggregation performed using the \verb|max| and \verb|mean| operator, respectively.

\begin{table*}[t!]
\small
\centering
\cprotect\caption{Average test AUROC, expressed as a percentage, for different state representations and behavior policy models in ADNI, RA, Sepsis, and COPD. History aggregation is performed using the \verb|mean| operator.}
\label{tab:results_mean}
\begin{tabular}{llllllllllllllllllllllllll}
\toprule
{} & \multicolumn{4}{c}{\textbf{ADNI}} & \multicolumn{3}{c}{\textbf{RA}} & \multicolumn{3}{c}{\textbf{Sepsis}} & \multicolumn{3}{c}{\textbf{COPD}} \\
\cmidrule(lr){2-5} \cmidrule(lr){6-8} \cmidrule(lr){9-11} \cmidrule(lr){12-14}  
{State} & {RS} & {LR} & {DT} & {MLP} & {LR} & {DT} & {MLP} & {LR} & {DT} & {MLP} & {LR} & {DT} & {MLP} \\
\midrule
$\bar{H}_t$ & 54.5 & 56.2 & 57.9 & 56.5 & 91.2 & 91.2 & 92.7 & 86.5 & 84.2 & 89.3 & 91.1 & 93.2 & 93.2 \\
$H_{(0)}, \bar{H}_t$ & 53.9 & 56.5 & 58.2 & 56.9 & 95.7 & 96.0 & 96.3 & 91.9 & 92.2 & 95.3 & 94.2 & 96.3 & 96.1 \\
$H_{(1)}, \bar{H}_t$ & 55.6 & 57.3 & 57.6 & 59.4 & 95.7 & 96.0 & 96.3 & 92.1 & 92.5 & 95.5 & 94.2 & 96.4 & 96.1 \\
$H_{(2)}, \bar{H}_t$ & 54.0 & 57.5 & 57.8 & 59.9 & 95.8 & 96.0 & 96.2 & 92.2 & 92.2 & 95.5 & 94.1 & 96.4 & 96.0 \\
\bottomrule
\end{tabular}
\end{table*}

\begin{table*}[t!]
\small
\centering
\cprotect\caption{Average test AUROC, expressed as a percentage, for different state representations and behavior policy models in ADNI, RA, Sepsis, and COPD. History aggregation is performed using the \verb|max| operator.}
\label{tab:results_max}
\begin{tabular}{llllllllllllllllllllllllll}
\toprule
{} & \multicolumn{4}{c}{\textbf{ADNI}} & \multicolumn{3}{c}{\textbf{RA}} & \multicolumn{3}{c}{\textbf{Sepsis}} & \multicolumn{3}{c}{\textbf{COPD}} \\
\cmidrule(lr){2-5} \cmidrule(lr){6-8} \cmidrule(lr){9-11} \cmidrule(lr){12-14}  
{State} & {RS} & {LR} & {DT} & {MLP} & {LR} & {DT} & {MLP} & {LR} & {DT} & {MLP} & {LR} & {DT} & {MLP} \\
\midrule
$\bar{H}_t$ & 54.1 & 55.6 & 54.6 & 55.6 & 90.8 & 89.8 & 91.9 & 84.3 & 83.5 & 88.7 & 90.3 & 92.6 & 92.9 \\
$H_{(0)}, \bar{H}_t$ & 54.9 & 57.2 & 54.4 & 57.2 & 95.9 & 96.0 & 96.6 & 91.9 & 92.8 & 95.2 & 94.8 & 96.3 & 96.5 \\
$H_{(1)}, \bar{H}_t$ & 55.1 & 58.3 & 58.2 & 59.5 & 95.9 & 95.9 & 96.6 & 92.3 & 92.1 & 95.5 & 94.8 & 96.4 & 96.6 \\
$H_{(2)}, \bar{H}_t$ & 55.8 & 58.1 & 58.0 & 60.4 & 95.9 & 96.0 & 96.6 & 92.4 & 92.3 & 95.6 & 94.7 & 96.5 & 96.5 \\
\bottomrule
\end{tabular}
\end{table*}

\begin{table*}[t]
\small
\centering
\cprotect\caption{Average test calibration error, expressed as a percentage, for different state representations and behavior policy models in ADNI, RA, Sepsis, and COPD. History aggregation is performed using the \verb|sum| operator. We report ECE for ADNI and SCE for RA, Sepsis, and COPD.}
\label{tab:cal_results}
\begin{tabular}{llllllllllllllllllllllllll}
\toprule
{} & \multicolumn{4}{c}{\textbf{ADNI}} & \multicolumn{3}{c}{\textbf{RA}} & \multicolumn{3}{c}{\textbf{Sepsis}} & \multicolumn{3}{c}{\textbf{COPD}} \\
\cmidrule(lr){2-5} \cmidrule(lr){6-8} \cmidrule(lr){9-11} \cmidrule(lr){12-14}  
{State} & {RS} & {LR} & {DT} & {MLP} & {LR} & {DT} & {MLP} & {LR} & {DT} & {MLP} & {LR} & {DT} & {MLP} \\
\midrule
$X_t$ & 8.6 & 1.9 & 3.5 & 2.2 & 1.1 & 1.3 & 1.6 & 0.2 & 0.3 & 0.2 & 0.4 & 0.4 & 0.5 \\
$A_{t-1}$ & 8.8 & 1.8 & 2.0 & 3.3 & 0.4 & 0.4 & 0.7 & 1.0 & 0.1 & 0.2 & 0.9 & 0.2 & 0.3 \\
$H_{(0)}$ & 8.2 & 2.3 & 4.0 & 2.8 & 0.6 & 0.4 & 0.8 & 0.6 & 0.1 & 0.2 & 0.6 & 0.2 & 0.3 \\
$\bar{H}_t$ & 6.5 & 2.8 & 2.7 & 3.9 & 3.5 & 1.3 & 1.1 & 0.4 & 0.2 & 0.2 & 0.6 & 0.5 & 0.4 \\
$H_{(0)}, \bar{H}_t$ & 10.1 & 2.6 & 2.6 & 3.6 & 0.9 & 0.5 & 0.9 & 0.6 & 0.1 & 0.2 & 0.5 & 0.2 & 0.4 \\
$H_{(1)}, \bar{H}_t$ & 7.6 & 2.7 & 3.3 & 2.9 & 1.0 & 0.5 & 0.8 & 0.5 & 0.1 & 0.2 & 0.5 & 0.2 & 0.4 \\
$H_{(2)}, \bar{H}_t$ & 7.9 & 2.7 & 3.1 & 2.4 & 1.0 & 0.5 & 1.0 & 0.5 & 0.1 & 0.2 & 0.5 & 0.2 & 0.4 \\
\addlinespace[1.5ex]
\midrule
\addlinespace[1.5ex]
{} & \multicolumn{4}{c}{\textbf{ADNI}} & \multicolumn{3}{c}{\textbf{RA}} & \multicolumn{3}{c}{\textbf{Sepsis}} & \multicolumn{3}{c}{\textbf{COPD}} \\
\cmidrule(lr){2-5} \cmidrule(lr){6-8} \cmidrule(lr){9-11} \cmidrule(lr){12-14}  
{State} & {CPR} & PSN & {RDT} & {RNN} & PSN & {RDT}& {RNN} & PSN & {RDT} & {RNN} & PSN & {RDT} & {RNN} \\
\midrule
$H_t$ & 2.2 & 3.8 & 2.4 & 4.4 & 1.1 & 4.0 & 0.8 & 0.5 & 0.9 & 0.2 & 0.6 & 0.3 & 0.4 \\
\bottomrule
\end{tabular}
\end{table*}

\begin{sidewaystable*}
\tiny
\centering
\cprotect\caption{Average test AUROC, expressed as a percentage, for the different states and behavior policy models in ADNI, RA, Sepsis, and COPD. History aggregation is performed using the \verb|sum| operator. The \SI{95}{\percent} confidence intervals are based on the bootstrap distribution of the average AUROC.}
\label{tab:models_full}
\begin{tabular}{lllllllll}
\toprule
{Data} & {State} & {RS} & {LR} & {DT} & {MLP} & {PSN} & {RDT} & {RNN} \\
\midrule
\multirow[c]{8}{*}{ADNI} & $X_t$ & \begin{tabular}[c]{@{}c@{}}54.2\\(51.4, 57.0)\end{tabular} & \begin{tabular}[c]{@{}c@{}}56.2\\(54.4, 57.4)\end{tabular} & \begin{tabular}[c]{@{}c@{}}53.9\\(52.1, 55.9)\end{tabular} & \begin{tabular}[c]{@{}c@{}}55.6\\(54.1, 57.0)\end{tabular} & - & - & - \\
& $A_{t-1}$ & \begin{tabular}[c]{@{}c@{}}52.0\\(51.1, 53.0)\end{tabular} & \begin{tabular}[c]{@{}c@{}}53.9\\(52.3, 55.6)\end{tabular} & \begin{tabular}[c]{@{}c@{}}53.8\\(52.4, 55.4)\end{tabular} & \begin{tabular}[c]{@{}c@{}}53.7\\(52.5, 55.4)\end{tabular} & - & - & - \\
& $H_{(0)}$ & \begin{tabular}[c]{@{}c@{}}53.4\\(51.1, 55.7)\end{tabular} & \begin{tabular}[c]{@{}c@{}}56.8\\(55.2, 58.1)\end{tabular} & \begin{tabular}[c]{@{}c@{}}54.3\\(52.7, 56.1)\end{tabular} & \begin{tabular}[c]{@{}c@{}}56.8\\(54.9, 58.8)\end{tabular} & - & - & - \\
& $\bar{H}_t$ & \begin{tabular}[c]{@{}c@{}}63.0\\(61.3, 64.7)\end{tabular} & \begin{tabular}[c]{@{}c@{}}64.4\\(62.8, 65.9)\end{tabular} & \begin{tabular}[c]{@{}c@{}}64.9\\(63.1, 66.7)\end{tabular} & \begin{tabular}[c]{@{}c@{}}64.1\\(62.2, 65.3)\end{tabular} & - & - & - \\
& $H_{(0)}, \bar{H}_t$ & \begin{tabular}[c]{@{}c@{}}63.7\\(62.9, 64.7)\end{tabular} & \begin{tabular}[c]{@{}c@{}}65.3\\(64.1, 66.4)\end{tabular} & \begin{tabular}[c]{@{}c@{}}65.0\\(63.1, 66.8)\end{tabular} & \begin{tabular}[c]{@{}c@{}}65.8\\(65.1, 66.5)\end{tabular} & - & - & - \\
& $H_{(1)}, \bar{H}_t$ & \begin{tabular}[c]{@{}c@{}}63.4\\(62.2, 64.6)\end{tabular} & \begin{tabular}[c]{@{}c@{}}65.6\\(64.4, 66.8)\end{tabular} & \begin{tabular}[c]{@{}c@{}}65.4\\(63.6, 67.1)\end{tabular} & \begin{tabular}[c]{@{}c@{}}66.0\\(65.3, 67.0)\end{tabular} & - & - & - \\
& $H_{(2)}, \bar{H}_t$ & \begin{tabular}[c]{@{}c@{}}62.9\\(61.7, 64.2)\end{tabular} & \begin{tabular}[c]{@{}c@{}}65.4\\(64.2, 66.7)\end{tabular} & \begin{tabular}[c]{@{}c@{}}65.3\\(63.6, 66.9)\end{tabular} & \begin{tabular}[c]{@{}c@{}}66.8\\(66.1, 67.6)\end{tabular} & - & - & - \\
& $H_t$ & - & - & - & - & \begin{tabular}[c]{@{}c@{}}66.7\\(65.6, 67.8)\end{tabular} & \begin{tabular}[c]{@{}c@{}}62.8\\(61.0, 64.7)\end{tabular} & \begin{tabular}[c]{@{}c@{}}68.0\\(67.2, 68.9)\end{tabular} \\
\midrule
\multirow[c]{8}{*}{RA} & $X_t$ & - & \begin{tabular}[c]{@{}c@{}}61.7\\(61.2, 62.4)\end{tabular} & \begin{tabular}[c]{@{}c@{}}58.8\\(58.1, 59.5)\end{tabular} & \begin{tabular}[c]{@{}c@{}}61.1\\(60.0, 62.1)\end{tabular} & - & - & - \\
& $A_{t-1}$ & - & \begin{tabular}[c]{@{}c@{}}94.7\\(94.4, 94.9)\end{tabular} & \begin{tabular}[c]{@{}c@{}}94.7\\(94.4, 94.9)\end{tabular} & \begin{tabular}[c]{@{}c@{}}94.7\\(94.4, 94.9)\end{tabular} & - & - & - \\
& $H_{(0)}$ & - & \begin{tabular}[c]{@{}c@{}}95.6\\(95.4, 95.7)\end{tabular} & \begin{tabular}[c]{@{}c@{}}95.7\\(95.5, 95.9)\end{tabular} & \begin{tabular}[c]{@{}c@{}}96.1\\(95.9, 96.2)\end{tabular} & - & - & - \\
& $\bar{H}_t$ & - & \begin{tabular}[c]{@{}c@{}}90.5\\(90.1, 90.9)\end{tabular} & \begin{tabular}[c]{@{}c@{}}92.0\\(91.4, 92.9)\end{tabular} & \begin{tabular}[c]{@{}c@{}}94.0\\(93.9, 94.2)\end{tabular} & - & - & - \\
& $H_{(0)}, \bar{H}_t$ & - & \begin{tabular}[c]{@{}c@{}}96.1\\(95.9, 96.2)\end{tabular} & \begin{tabular}[c]{@{}c@{}}96.5\\(96.3, 96.6)\end{tabular} & \begin{tabular}[c]{@{}c@{}}96.9\\(96.7, 97.0)\end{tabular} & - & - & - \\
& $H_{(1)}, \bar{H}_t$ & - & \begin{tabular}[c]{@{}c@{}}96.0\\(95.9, 96.1)\end{tabular} & \begin{tabular}[c]{@{}c@{}}96.4\\(96.2, 96.6)\end{tabular} & \begin{tabular}[c]{@{}c@{}}96.9\\(96.7, 97.0)\end{tabular} & - & - & - \\
& $H_{(2)}, \bar{H}_t$ & - & \begin{tabular}[c]{@{}c@{}}96.0\\(95.8, 96.1)\end{tabular} & \begin{tabular}[c]{@{}c@{}}96.4\\(96.1, 96.6)\end{tabular} & \begin{tabular}[c]{@{}c@{}}96.7\\(96.6, 96.9)\end{tabular} & - & - & - \\
& $H_t$ & - & - & - & - & \begin{tabular}[c]{@{}c@{}}96.2\\(96.1, 96.4)\end{tabular} & \begin{tabular}[c]{@{}c@{}}90.0\\(85.9, 94.1)\end{tabular} & \begin{tabular}[c]{@{}c@{}}96.8\\(96.7, 97.0)\end{tabular} \\
\midrule
\multirow[c]{8}{*}{Sepsis} & $X_t$ & - & \begin{tabular}[c]{@{}c@{}}82.1\\(81.8, 82.4)\end{tabular} & \begin{tabular}[c]{@{}c@{}}78.2\\(77.3, 78.9)\end{tabular} & \begin{tabular}[c]{@{}c@{}}84.1\\(83.9, 84.3)\end{tabular} & - & - & - \\
& $A_{t-1}$ & - & \begin{tabular}[c]{@{}c@{}}88.0\\(87.9, 88.1)\end{tabular} & \begin{tabular}[c]{@{}c@{}}90.6\\(90.1, 91.0)\end{tabular} & \begin{tabular}[c]{@{}c@{}}91.1\\(91.0, 91.2)\end{tabular} & - & - & - \\
& $H_{(0)}$ & - & \begin{tabular}[c]{@{}c@{}}91.3\\(91.2, 91.4)\end{tabular} & \begin{tabular}[c]{@{}c@{}}92.1\\(90.9, 92.9)\end{tabular} & \begin{tabular}[c]{@{}c@{}}94.7\\(94.6, 94.8)\end{tabular} & - & - & - \\
& $\bar{H}_t$ & - & \begin{tabular}[c]{@{}c@{}}84.6\\(84.4, 84.8)\end{tabular} & \begin{tabular}[c]{@{}c@{}}85.2\\(84.4, 86.0)\end{tabular} & \begin{tabular}[c]{@{}c@{}}89.1\\(89.0, 89.2)\end{tabular} & - & - & - \\
& $H_{(0)}, \bar{H}_t$ & - & \begin{tabular}[c]{@{}c@{}}91.9\\(91.8, 92.0)\end{tabular} & \begin{tabular}[c]{@{}c@{}}92.3\\(91.4, 92.9)\end{tabular} & \begin{tabular}[c]{@{}c@{}}95.3\\(95.2, 95.3)\end{tabular} & - & - & - \\
& $H_{(1)}, \bar{H}_t$ & - & \begin{tabular}[c]{@{}c@{}}92.2\\(92.1, 92.3)\end{tabular} & \begin{tabular}[c]{@{}c@{}}92.5\\(92.2, 92.9)\end{tabular} & \begin{tabular}[c]{@{}c@{}}95.5\\(95.4, 95.5)\end{tabular} & - & - & - \\
& $H_{(2)}, \bar{H}_t$ & - & \begin{tabular}[c]{@{}c@{}}92.3\\(92.2, 92.4)\end{tabular} & \begin{tabular}[c]{@{}c@{}}92.6\\(92.2, 92.9)\end{tabular} & \begin{tabular}[c]{@{}c@{}}95.5\\(95.4, 95.6)\end{tabular} & - & - & - \\
& $H_t$ & - & - & - & - & \begin{tabular}[c]{@{}c@{}}94.9\\(94.7, 95.1)\end{tabular} & \begin{tabular}[c]{@{}c@{}}77.0\\(69.5, 85.5)\end{tabular} & \begin{tabular}[c]{@{}c@{}}95.7\\(95.7, 95.8)\end{tabular} \\
\midrule
\multirow[c]{8}{*}{COPD} & $X_t$ & - & \begin{tabular}[c]{@{}c@{}}77.9\\(77.4, 78.3)\end{tabular} & \begin{tabular}[c]{@{}c@{}}74.7\\(72.9, 75.9)\end{tabular} & \begin{tabular}[c]{@{}c@{}}78.7\\(78.1, 79.4)\end{tabular} & - & - & - \\
& $A_{t-1}$ & - & \begin{tabular}[c]{@{}c@{}}92.9\\(92.5, 93.2)\end{tabular} & \begin{tabular}[c]{@{}c@{}}95.0\\(94.6, 95.2)\end{tabular} & \begin{tabular}[c]{@{}c@{}}94.9\\(94.9, 95.0)\end{tabular} & - & - & - \\
& $H_{(0)}$ & - & \begin{tabular}[c]{@{}c@{}}94.0\\(93.8, 94.2)\end{tabular} & \begin{tabular}[c]{@{}c@{}}96.0\\(95.7, 96.3)\end{tabular} & \begin{tabular}[c]{@{}c@{}}95.5\\(95.2, 95.7)\end{tabular} & - & - & - \\
& $\bar{H}_t$ & - & \begin{tabular}[c]{@{}c@{}}91.1\\(90.8, 91.4)\end{tabular} & \begin{tabular}[c]{@{}c@{}}89.3\\(86.7, 91.4)\end{tabular} & \begin{tabular}[c]{@{}c@{}}93.6\\(93.2, 93.9)\end{tabular} & - & - & - \\
& $H_{(0)}, \bar{H}_t$ & - & \begin{tabular}[c]{@{}c@{}}94.7\\(94.6, 94.9)\end{tabular} & \begin{tabular}[c]{@{}c@{}}96.7\\(96.2, 97.0)\end{tabular} & \begin{tabular}[c]{@{}c@{}}96.3\\(96.1, 96.5)\end{tabular} & - & - & - \\
& $H_{(1)}, \bar{H}_t$ & - & \begin{tabular}[c]{@{}c@{}}94.7\\(94.5, 94.9)\end{tabular} & \begin{tabular}[c]{@{}c@{}}96.8\\(96.4, 97.0)\end{tabular} & \begin{tabular}[c]{@{}c@{}}96.4\\(96.2, 96.6)\end{tabular} & - & - & - \\
& $H_{(2)}, \bar{H}_t$ & - & \begin{tabular}[c]{@{}c@{}}94.7\\(94.5, 94.8)\end{tabular} & \begin{tabular}[c]{@{}c@{}}96.8\\(96.4, 97.0)\end{tabular} & \begin{tabular}[c]{@{}c@{}}96.3\\(96.1, 96.5)\end{tabular} & - & - & - \\
& $H_t$ & - & - & - & - & \begin{tabular}[c]{@{}c@{}}96.2\\(96.0, 96.5)\end{tabular} & \begin{tabular}[c]{@{}c@{}}81.9\\(76.1, 86.2)\end{tabular} & \begin{tabular}[c]{@{}c@{}}96.5\\(96.3, 96.6)\end{tabular} \\
\bottomrule
\end{tabular}
\end{sidewaystable*}

\end{document}